\documentclass{article}

\usepackage[utf8]{inputenc}
\usepackage[T1]{fontenc}
\usepackage{amsmath}
\usepackage{iclr2025_conference}
\usepackage{booktabs}
\usepackage{array}
\usepackage{tabularx}
\usepackage{longtable}
\usepackage{multirow}
\usepackage{makecell}
\usepackage[font=small,labelfont=bf]{caption}
\usepackage[section]{placeins}
\usepackage{tikz}
\usepackage{listings}

\definecolor{errbarblue}{RGB}{220,37,42}
\definecolor{errbargray}{RGB}{235,235,235}
\newcommand{\scorebarerr}[2]{%
\begin{tikzpicture}[baseline=(n.base)]
  \node[inner sep=0pt, outer sep=0pt, anchor=west] (n) at (0,0) {%
    \begin{tikzpicture}[x=1mm,y=1mm,baseline=(m.base)]
      \fill[errbargray] (0,0) rectangle (15,2.2);
      \pgfmathsetmacro{\w}{15*#1/100}
      \fill[errbarblue] (0,0) rectangle (\w,2.2);
      \node[anchor=west, font=\scriptsize] (m) at (16.2,1.1) {#2};
    \end{tikzpicture}%
  };
\end{tikzpicture}%
}

\newcommand{\evolve}{\textsc{EvolveScaler}}
\newcommand{\ie}{information evolution}
\newcommand{\IE}{Information Evolution}

\title{\evolve{}: Synthesizing Information-Evolution Contexts via Executable State Machines and Natural-Language Rendering}
\author{
Ziliang Zhao$^{1,2,*}$, Zenan Xu$^{2,*}$, Shuting Wang$^1$, Zhao Wang$^{1,2}$, Bowen Cao$^{1,3}$, Minda Hu$^{1,3}$, Lincheng Li$^2$, Pluto Zhou$^{2,\dagger}$, Zhicheng Dou$^{1,\dagger}$  \\
\vspace{1mm}
\textbf{$^1$Gaoling School of Artificial Intelligence, Renmin University of China} \\
\textbf{$^2$LLM Department, Hunyuan Team, Tencent} \quad \textbf{$^3$The Chinese University of Hong Kong} \\
\vspace{1mm}
{\small $^*$Equal contribution.\quad $^\dagger$Corresponding author.}
}
\begin{document}
\maketitle

\begin{abstract}

In persistent interactions, a long context can encode an evolving process rather than a fixed record of facts. Later events can revise or revoke earlier records, changing which information remains valid and what conclusions follow. We refer to this setting as \emph{\ie{}} (IE). Solving an IE task requires identifying valid records, applying updates in order, and reconstructing the query-relevant state from the event history. Constructing reliable IE data is difficult because conventional pipelines generate long-form text before deriving supervision, leaving state transitions implicit and answers difficult to verify. To address this problem, we introduce \evolve{}, a code-driven framework that defines information evolution in code before rendering it as natural language. Human-authored operational specifications define how events alter state and which records remain valid. They also specify difficulty controls and executable answer logic. A strong LLM synthesizes a self-contained simulator from each specification. Executing a validated simulator produces natural-language, multi-turn event histories, while deterministic replay computes reference answers and atomic checklists. The LLM proposes the executable mechanism, while the replayed program state determines the supervision. We instantiate \evolve{} with $117$ task prototypes and $159$ final-question operators. Five controlled difficulty levels jointly increase trajectory scale and evolution complexity, spanning approximately $7$ to $1{,}200$ events per instance. The resulting resource contains approximately $35{,}100$ training examples and $585$ validated evaluation instances. Using the evaluation instances, we benchmark frontier and open-source models across the five difficulty levels. On the \texttt{very\_long} tier, the strongest model achieves an avg@5 of $59.3\%$, while six models have avg@5 scores below $10\%$. To test the training value of the generated data, we train an internal A3B model on $6{,}000$ \evolve{} examples. The trained model outperforms its base checkpoint on all eight independently constructed out-of-distribution benchmarks, with an average gain of $5.25$ points. These results show that code-driven IE synthesis supports both diagnostic evaluation and transferable training supervision.\footnote{Project page: \url{https://tencent-hunyuan.github.io/evolve-scaler/}. Correspondence to: zhaoziliang@ruc.edu.cn}

\end{abstract}

% ===========================================================================
\section{Introduction}
% ===========================================================================

As large language models (LLMs) are increasingly expected to support tasks that unfold over extended interactions, they must reason over information introduced as those interactions progress~\citep{dou2026cllife,dou2026cl}. However, many long-context and multi-turn task formulations still treat the context as a \textbf{relatively static snapshot} whose task-relevant information remains largely stable while the task is being solved. This view breaks down in persistent settings because information does not merely accumulate. A later update may revise an earlier record and invalidate conclusions drawn from it. Even when the complete interaction history is presented as a fixed input sequence, the context remains \textbf{semantically dynamic}. Its interpretation depends on how successive updates transform the state established so far. Such a context is not merely a longer document to read but an event history to replay. A model must therefore recover the state required by the current query from that ordered history.

\begin{figure}[tp]
\centering
\includegraphics[width=\textwidth]{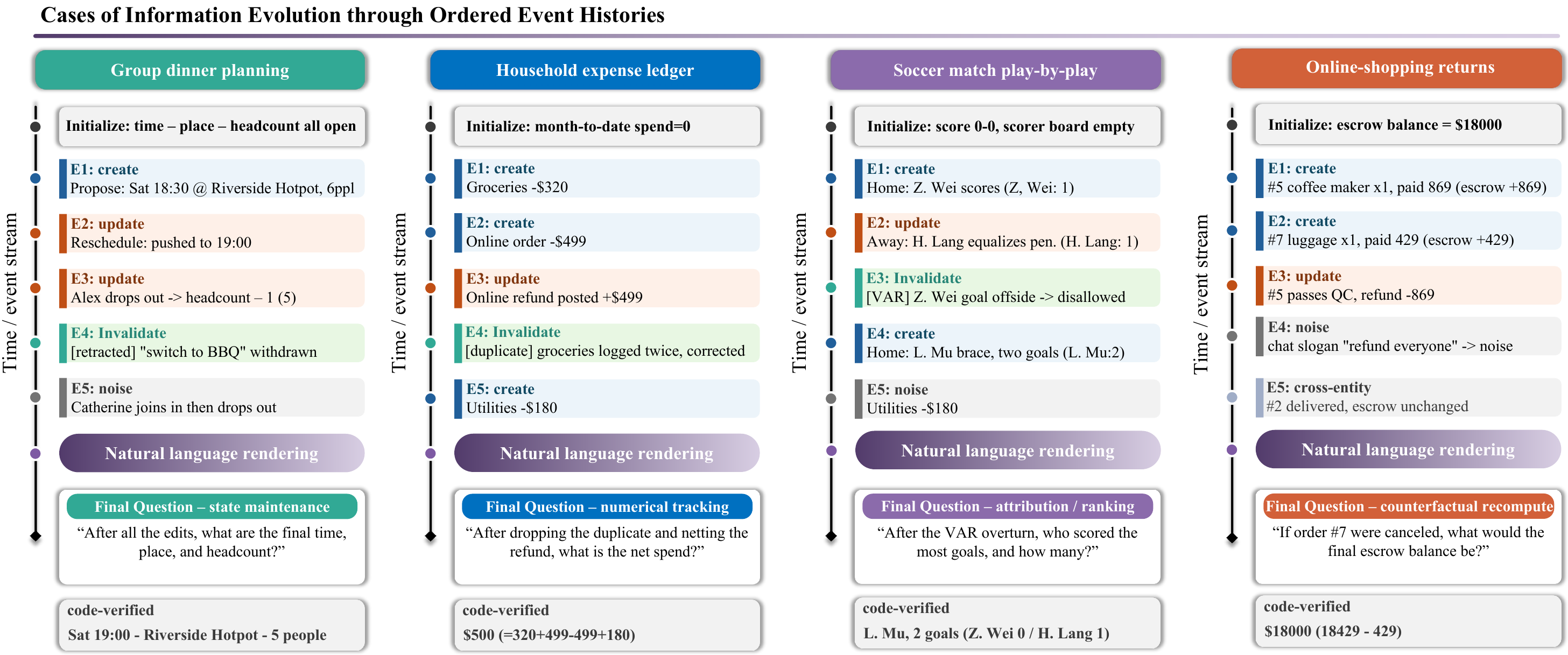}
\caption{Four information-evolution scenarios drawn from task prototypes in group dinner planning, a household expense ledger, soccer match play-by-play, and online shopping returns. Across the examples, state updates are interleaved with invalid, state-preserving, and cross-entity records. The final questions probe different views of the reconstructed state and span varying difficulty levels; each answer is computed by deterministic replay.}
\label{fig:examples}
\end{figure}

We refer to this setting as \textbf{\IE{}} (IE) and to the resulting context as an information-evolution context (IEC). As Figure~\ref{fig:examples} illustrates across four domains, different final questions request different views of the evolving state, but all require the model to resolve record validity and apply updates in order. IE therefore adds a source of difficulty beyond context length or turn count alone. Even when two interactions have comparable context lengths and the same number of turns, their difficulty can differ sharply if one merely accumulates information that remains valid while the other repeatedly revises or invalidates earlier records. Long-context benchmarks have substantially advanced retrieval and reasoning over extended inputs, but many still center on static documents or synthetic retrieval targets~\citep{yang2025marsbench,bai2023longbench,hsieh2024ruler,liu2023lost,shaham2023zeroscrolls,an2023eval}. Multi-turn benchmarks primarily examine instruction retention and cross-turn coherence~\citep{zheng2023judging,deshpande2025multichallenge,laban2025lostconversation}. Although some tasks include corrections or versioned edits~\citep{deshpande2025multichallenge}, update semantics are rarely formalized as an explicit and controllable part of the task. Existing evaluations therefore do not yet isolate IE systematically across domains and difficulty levels.

Reliable evaluation and effective training for IE require data built from an explicit specification of how information evolves, with supervision derived from the same process. Conventional synthesis pipelines instead generate instructions or conversations directly in natural language~\citep{wang2022selfinstruct,xu2023wizardlm,ding2023ultrachat}. For IE, this text-first workflow leaves transition semantics and answer logic implicit, making consistency and correctness increasingly difficult to guarantee as trajectories grow. We therefore reverse the pipeline by \textbf{defining information evolution in code before rendering it as natural language}. In \evolve{}, a human-authored operational specification defines the state-transition system, record-validity rules, difficulty controls, and executable answer logic for each task prototype. A frontier LLM then synthesizes a self-contained simulator under a fixed generation contract. Only simulators that preserve the annotated constraints and reproduce the same state under deterministic replay are retained. Accepted simulators map each state transition to a corresponding natural-language description according to the specified rendering rules, thereby producing multi-turn event histories, while the underlying program state yields the reference answer and atomic checklist. The LLM thus proposes the executable mechanism, but the program state determines the supervision. This design combines long and mutable contexts, dense and controllable event streams, and deterministically verifiable supervision. Explicit controls over trajectory scale and evolution complexity allow each simulator to span training-oriented warm-ups and evaluation instances that challenge frontier models.

We instantiate the framework at scale with $117$ task prototypes authored by more than twenty annotators and spanning twelve themes. The synthesis and validation pipeline yields $117$ accepted simulators supporting $159$ distinct final-question operators. Sampling these simulators across five tiers that jointly scale trajectory length and evolution complexity produces approximately $35{,}100$ training examples. We generate the held-out pool with a random seed disjoint from training and retain $585$ validated examples for model evaluation. We evaluate the resulting data in two intended roles. For evaluation, deterministically computed answers provide a trustworthy basis for comparing information-evolution capability across $14$ frontier and open-source LLMs. The median avg@5 falls from $71.2$ on the shortest tier to $11.3$ on the longest, revealing a sharp loss of reliability as evolution histories grow. For training, independently sampled traces provide scalable code-derived supervision. Continued training on $6{,}000$ examples improves performance on all eight independently constructed out-of-distribution benchmarks and raises their average score by $5.25$ points. Because these benchmarks are absent from the training data, the breadth of the gains supports the transfer of state-tracking and evidence-aggregation capabilities to out-of-distribution settings beyond the \evolve{} task format.

Our contributions are threefold.
\begin{itemize}
    \setlength{\topsep}{0.25em}
    \setlength{\itemsep}{0.2em}
    \setlength{\parsep}{0pt}

    \item[(i)] We formulate \IE{} (IE) as a distinct reasoning setting not captured by context length or turn count alone, and introduce the information-evolution context (IEC), in which models must reconstruct query-relevant state from an ordered event history whose updates can change the validity and consequences of prior information.

    \item[(ii)] We introduce \evolve{}, a code-driven synthesis framework that defines information evolution in code before rendering it as natural language. It uses an LLM to synthesize executable simulators from human-authored operational specifications. These simulators preserve the specified state and validity semantics while generating long, mutable histories with controllable difficulty and deterministic supervision.

    \item[(iii)] We instantiate the framework with $117$ task prototypes and $159$ final-question operators, and establish the value of the resulting data in its two intended roles. As an evaluation suite, deterministically computed answers support trustworthy comparison across $14$ LLMs and reveal a sharp decline in reliability as trajectory scale and evolution complexity increase. As a training source, continued training improves performance on all eight independently constructed out-of-distribution benchmarks, providing evidence that state-tracking and evidence-aggregation capabilities transfer beyond the \evolve{} task format.
\end{itemize}

% ===========================================================================
\section{Related Work}

\subsection{Long-Context and Multi-Turn Evaluation}

Long-context benchmarks study whether models can retrieve, integrate, and reason over evidence distributed across extended inputs. LongBench, SCROLLS, ZeroSCROLLS, L-Eval, LooGLE, InfiniteBench, and HELMET span single- and multi-document question answering, summarization, and other long-text understanding tasks across a range of context lengths~\citep{bai2023longbench,shaham2022scrolls,shaham2023zeroscrolls,an2023eval,li2023loogle,zhang2024infinitebench}. Building on needle-in-a-haystack tests, RULER introduces configurable retrieval tasks together with multi-hop tracing and aggregation~\citep{hsieh2024ruler}. BABILong embeds fact chaining, induction, deduction, counting, and related reasoning tasks within extremely long documents~\citep{kuratov2024babilong}, while LongBench v2 focuses on realistic long-context problems that require deeper understanding and reasoning~\citep{bai2025longbenchv2}. Lost in the Middle further shows that models may fail to use relevant evidence when it appears away from favorable context positions~\citep{liu2023lost}. Together, these benchmarks show that accepting a long input does not guarantee reliable use of the information it contains. In many of these settings, however, the context remains fixed during inference. Later records do not typically revise the validity or consequences of earlier evidence.

Multi-turn benchmarks examine failures that emerge when information and instructions are distributed across an interaction. MT-Bench and MT-Bench-101 evaluate multi-turn chat and instruction-following behavior, while Chatbot Arena provides large-scale pairwise preference comparisons of chat models~\citep{zheng2023judging,bai2024mtbench101}. IFEval evaluates compliance with programmatically verifiable instructions, and Multi-IF extends this setting to multi-turn and multilingual conversations~\citep{zhou2023ifeval,he2024multiif}. MultiChallenge covers instruction retention, inference memory, versioned editing, and self-coherence~\citep{deshpande2025multichallenge}. Recent work also shows that tasks that are straightforward when fully specified in one turn can become substantially harder when the same information is revealed over a longer interaction~\citep{laban2025lostconversation}. These evaluations capture instruction retention, cross-turn dependence, and some forms of revision. Even when revisions are present, the underlying update semantics are not generally represented as an executable state process with explicit controls over record validity and difficulty. \evolve{} focuses on this dimension by treating ordered state changes and record-validity rules as part of the task specification.

\subsection{State Tracking, Memory, and Dynamic Environments}

Dialogue state tracking is closely related to \ie{} because it represents an interaction through successive changes to a dialogue state. DSTC, MultiWOZ, and Schema-Guided Dialogue model updates to slot-based states across turns~\citep{williams2013dstc,budzianowski2018multiwoz,rastogi2019sgd}. LoCoMo and LongMemEval evaluate long-term conversational memory across extended or multi-session interactions~\citep{maharana2024locomo,wu2024longmemeval}. MemGPT studies how memory can be managed beyond the immediate context window~\citep{packer2023memgpt}, while entity-state probing directly tests whether models can infer an entity's final state after a sequence of state-changing operations~\citep{kim2023entitytracking}. \evolve{} builds on these concerns but targets histories in which records may be revised, invalidated, backfilled, or mixed with state-preserving noise. It also allows different final questions to request different views of the reconstructed state from the same underlying process, a capability that previous studies emphasize less.

MARS-Bench is particularly close in its use of event streams. It constructs multi-turn evaluation scenarios from real-world sports play-by-play data and uses these event histories to test complex dependencies across turns~\citep{yang2025marsbench}. Its use of real commentary provides naturally occurring event streams. \evolve{} takes a complementary programmatic approach by encoding the event system itself, allowing trajectory scale, record validity, noise, and question difficulty to be controlled while reference answers remain deterministic. WebArena and $\tau$-bench evaluate agents that interact with external web or database-backed environments over multiple steps~\citep{zhou2023webarena,yao2024taubench}. Other agent benchmarks cover software-issue resolution, function calling, multi-environment evaluation, and general assistant tasks~\citep{jimenez2023swebench,patil2025bfcl,liu2023agentbench,mialon2023gaia}. In interactive environment benchmarks, state management is assessed together with tool use, planning, and action selection. \evolve{} instead provides the event history as context and focuses on reconstructing and querying the state encoded by that history.

\subsection{Synthetic Data and Executable Supervision}

LLM-generated synthetic data has become an important source of training examples. Self-Instruct, WizardLM, and UltraChat show that model-generated instructions and conversations can improve instruction following and dialogue capabilities~\citep{wang2022selfinstruct,xu2023wizardlm,ding2023ultrachat}. Related work extends this paradigm through instruction tuning, distillation, textbook-style synthesis, prompt-based data generation, and self-play~\citep{taori2023alpaca,mukherjee2023orca,gunasekar2023textbooks,xu2024magpie,chen2024spin,liu2024synthsurvey}. These approaches generally construct examples and supervision directly in natural language. They do not typically require an independent executable process that verifies each state transition and recomputes the target answer.

Programmatic datasets and simulator-based environments provide stronger structural guarantees. bAbI, CLEVR, SCAN, PrOntoQA, and ProofWriter derive examples from synthetic worlds, functional programs, or formal rule systems~\citep{weston2015babi,johnson2017clevr,lake2018scan,saparov2023prontoqa,tafjord2021proofwriter}. TextWorld and ALFWorld provide executable interactive environments for training and evaluating agents~\citep{cote2018textworld,shridhar2020alfworld}. These works primarily target text understanding, compositional reasoning, visual question answering, formal inference, or agent actions. LongAlign and work on data engineering for 128K contexts introduce data and training recipes that improve the use of extended input sequences~\citep{bai2024longalign,fu2024dataeng}. Their primary goal is to extend long-context modeling rather than to model how later records alter the validity of earlier information. \evolve{} assigns different responsibilities across the synthesis process. Human-authored operational specifications define state-transition and record-validity semantics together with difficulty controls and answer logic. An LLM synthesizes the executable simulator, while program execution and deterministic replay determine the reference answer and atomic checklist. This design produces natural-language information-evolution histories whose difficulty is controllable and whose supervision remains independently verifiable.

% ===========================================================================
\section{The \evolve{} Framework}
% ===========================================================================

\begin{figure}[tp]
\centering
\includegraphics[width=\textwidth]{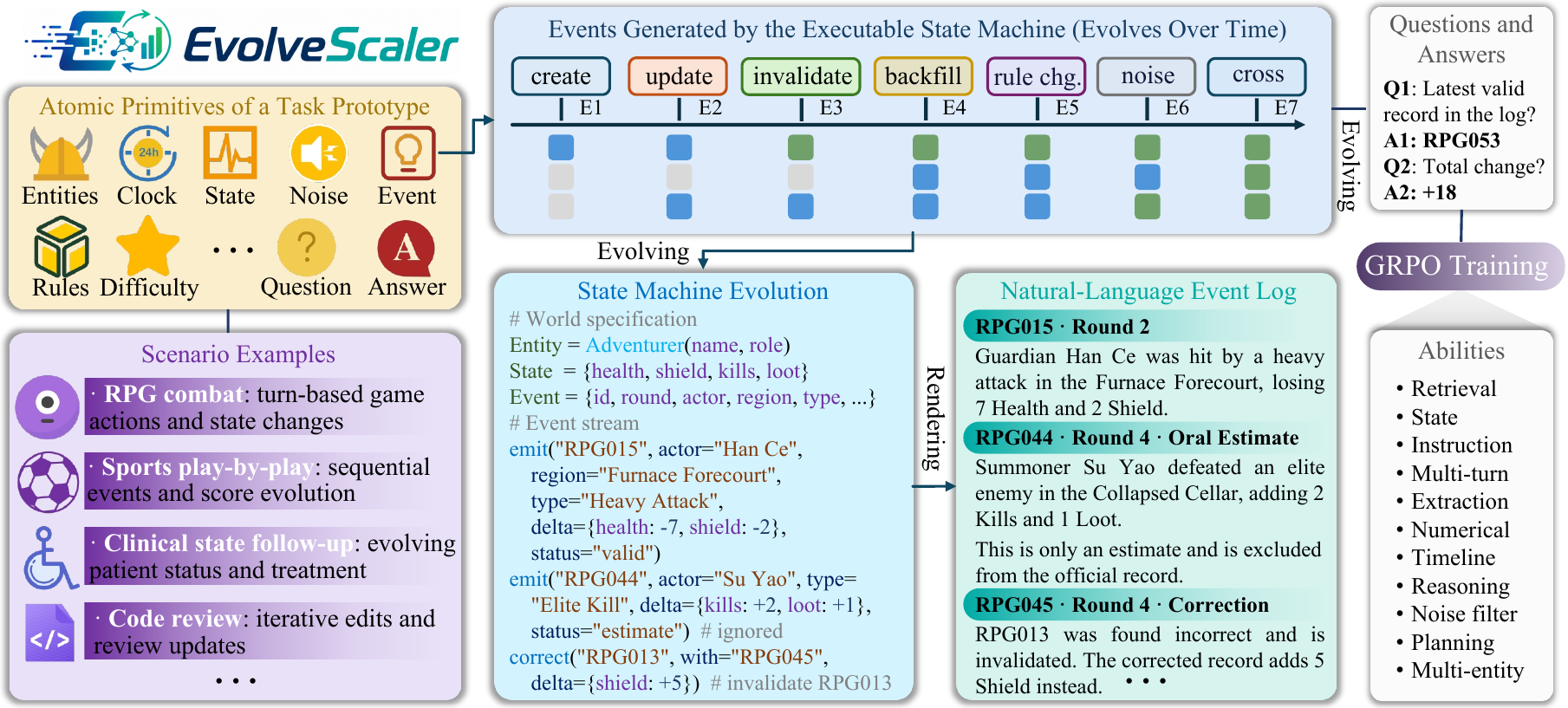}
\caption{\textbf{The \evolve{} framework.} Starting from atomic task primitives, \evolve{} synthesizes executable state machines that simulate evolving entities, states, and events over time, renders these structured trajectories into diverse natural-language contexts, and automatically produces verifiable question--answer pairs for training and evaluation.}
\label{fig:overview}
\end{figure}

\evolve{} follows a code-first design that separates task semantics, executable realization, and supervision. Each task prototype is formalized as a human-authored operational specification $\Pi$, which defines how events alter the state, which records remain valid, how difficulty is controlled, and how answers are computed. Conditioned on $\Pi$, a strong LLM synthesizes a self-contained simulator $P_{\Pi}$ under a fixed generation contract. The LLM thus proposes an executable mechanism rather than directly authoring an unverifiable long-form sample. The specification governs the task semantics and difficulty, while the resulting program state determines the supervision. Figure~\ref{fig:overview} summarizes these roles and their interaction.

A validated simulator represents a family of IE instances rather than a single fixed sample. Varying the random seed, question operator, and difficulty setting changes the realized event history and the requested view of its reconstructed state without changing the underlying task semantics. Each execution renders the event history as a natural-language, multi-turn context and samples a final question. Deterministic replay of the same history yields the reference answer and atomic checklist. The central technical challenge is therefore to synthesize diverse executable simulators while ensuring that their transitions and supervision remain faithful to the operational specification. We next describe how \evolve{} generates and verifies $P_{\Pi}$.

\subsection{Code-Driven Simulator Synthesis and Deterministic Supervision}
\label{sec:simulator-synthesis}

Directly asking an LLM to generate both a long IE trajectory and its target answer makes correctness difficult to verify. The model must preserve the consequences of every update while deriving the answer from the same generated text. As trajectories grow, a local inconsistency can propagate through later updates, while a model-generated answer provides no independent check that the underlying state has been tracked correctly. To address these issues, \evolve{} shifts the role of the LLM from directly generating trajectories and labels to synthesizing an executable simulator, while program execution determines the supervision.

\paragraph{Executable simulator synthesis.}
Given an operational specification $\Pi$, we prompt a strong LLM to generate a \emph{self-contained Python simulator} $P_{\Pi}$. Rather than directly authoring a long-form instance, the LLM implements the operational semantics encoded by $\Pi$ under a fixed generation contract. The contract specifies the complete program interface from state initialization and event execution to natural-language rendering and answer computation. Crucially, hard constraints in $\Pi$ are compiled into transition guards and runtime assertions instead of being left as prose for the LLM to follow implicitly. For example, an inventory-decrement event is legal only when sufficient stock remains. A refund must reference a valid earlier payment and cannot exceed its unsettled amount. The simulator therefore enforces legal state evolution during execution rather than attempting to recover consistency from the rendered text afterward.

\paragraph{Program-level verification and deterministic supervision.}
A generated program may run successfully yet still fail to implement $\Pi$ faithfully. We retain $P_{\Pi}$ only if it executes successfully across sampled random seeds, preserves all annotated invariants, reproduces the same state under deterministic replay, and produces an atomic checklist consistent with its reference answer. More than twenty annotators additionally inspect the logical consistency of the synthesized data; this audit finds zero errors in a random sample of $117$ instances. For an accepted simulator and random seed $\omega$, the execution contract is
\begin{equation}
\operatorname{Exec}(P_{\Pi};\omega)
=
\bigl(
C_{\omega},
q_{\omega},
a_{\omega}^{\star},
\chi_{\omega}
\bigr),
\qquad
a_{\omega}^{\star}
=
G_{q_{\omega}}\!\left(
\operatorname{Replay}_{\Pi}(e_{1:T};\omega)
\right),
\label{eq:program-contract}
\end{equation}
where $e_{1:T}$ is the generated event history, $C_{\omega}$ is its rendered natural-language context, $q_{\omega}$ is the sampled final question, $a_{\omega}^{\star}$ is the reference answer, and $\chi_{\omega}$ is the corresponding atomic checklist. The executable answer program $G_{q_{\omega}}$ applies the sampled question operator to the state reconstructed through replay. The checklist is derived from the same replayed state and must remain consistent with the reference answer. For counterfactual questions, the replay operator removes the designated subset of events before recomputing the requested state view.

This design separates model-assisted synthesis from ground-truth computation. The LLM determines how the operational specification is realized as an executable program, but the replayed program state determines the reference answer and checklist. Supervision therefore does not rely on a post hoc model judgment over the rendered text, and natural-language variation does not change the underlying answer semantics. The next subsection defines the operational specification $\Pi$ that constrains simulator synthesis and replay.

\subsection{Executable Specifications for Information Evolution}
\label{sec:ie-specification}

The synthesis and verification procedure above is meaningful only if the intended task semantics are defined independently of the LLM-generated simulator. \evolve{} establishes this semantic reference through a human-authored operational specification $\Pi$, which turns each task prototype into an executable contract. Annotators begin from prototypes drawn from process-intensive settings in which records are routinely added, revised, or invalidated. The specification serves two complementary roles. It defines the evolving world that every simulator must preserve, and it defines how that world is queried and scaled into concrete instances. Table~\ref{tab:atoms} summarizes its constituent fields, while Appendix~\ref{app:domains} provides the complete inventory of task prototypes.

\paragraph{Evolution semantics.}
The first role of $\Pi$ is to define how information changes over time. It specifies the entities and temporal structure of the process, the admissible event schemas, the state maintained by the simulator, and the rules governing legal transitions, record validity, and state-preserving noise. For an event $a_t$ with parameters $\theta_t$, a state update is admissible only if it satisfies every domain invariant:
\begin{equation}
s_t
=
\delta_{a_t}(s_{t-1};\theta_t),
\qquad
g_r(s_t) \leq 0,\ \forall r,
\label{eq:invariants}
\end{equation}
where each function $g_r$ encodes a domain-specific requirement, such as non-negative inventory, refunds not exceeding prior payments, capacity limits, or precedence constraints. Transition invariants determine whether a proposed event can legally alter the state. Record-validity rules address a different question by determining whether a record shown in the context should contribute during replay. A plausible record may still be excluded because it is duplicated, retracted, superseded, or otherwise invalid, while noise records leave the tracked state unchanged. Together, these rules ensure that every generated event has a well-defined state effect and every invalid record has an explicit exclusion rule. The running state is never exposed directly in the rendered context. Individual records reveal only local changes, such as ``stock $+2$'', so the evaluated model must identify which records apply and reconstruct the required state from the event history.

\paragraph{Question and difficulty controls.}
Once the evolution semantics are fixed, the second role of $\Pi$ is to define how the underlying process becomes a concrete training or evaluation instance. The difficulty controls determine the trajectory length and evolution complexity. A question operator $q$ selects the requested view of the reconstructed state, and an executable answer program $G_q$ computes the corresponding reference answer. Varying the difficulty changes how much information evolution must be processed, while varying $q$ changes what must be recovered from the same underlying process. The same task prototype can therefore support multiple context scales and reasoning demands without changing its state-transition semantics. The next subsection describes how these controls are instantiated together with natural-language rendering to construct the final instances.

\subsection{Controllable Instance Generation and Corpus Construction}
\label{sec:data}

With the evolution semantics fixed, \evolve{} constructs concrete IE instances at three levels. The question operator and difficulty setting determine which state view must be recovered and how demanding the recovery is. Multi-turn context construction determines how the evolving history is presented to the model. Repeated sampling across simulators and difficulty tiers then produces the final training and held-out corpora.

\begin{table}[tp]
\centering
\footnotesize
\setlength{\tabcolsep}{4pt}
\renewcommand{\arraystretch}{1.12}
\caption{\textbf{Atomic primitives of a task prototype.} Examples are drawn from a group-trip shared-expense chat.
The last column reports the corresponding inventory measured across all $117$ prototypes.}
\label{tab:atoms}
\begin{tabularx}{\textwidth}{@{}>{\raggedright\arraybackslash}p{1.95cm}>{\raggedright\arraybackslash}p{2.25cm}>{\raggedright\arraybackslash}X>{\raggedright\arraybackslash}p{3.3cm}@{}}
\toprule
\textbf{Primitive} & \textbf{What it specifies} & \textbf{Concrete example (a group-trip expense chat)} & \textbf{In the corpus} \\
\midrule
Entities & typed actors and objects & travelers (Ann, Bob, Carl, Dan); expense items (flights, hotel, car rental, meals, tickets); categories (transport, lodging, food) & $3{,}410$ named entities; median $28$ per prototype (range $19$--$52$) \\
\cmidrule(lr){1-4}
Clock / block & temporal granularity & a trip day or planning round (e.g., ``Day~2'', ``round~5''), each containing a batch of newly posted messages & $20$ distinct clock schemes; $3$--$55$ blocks per sample \\
\cmidrule(lr){1-4}
Event schemas & admissible updates & add or revise an expense; cancel a booking and issue a refund; add or remove a traveler; confirm a tentative quote & $928$ schemas, median $8$ each (range $5$--$15$), realized by $2{,}784$ phrasings \\
\cmidrule(lr){1-4}
State variables & quantities maintained over time & total trip cost; each traveler's paid amount, liability, and balance; category-level spending; confirmed deposits and refundable amounts & $549$ tracked quantities, median $5$ each (range $4$--$7$) \\
\cmidrule(lr){1-4}
Transition constraints & legal state transitions & a refund cannot exceed the corresponding payment; only current members share a new expense; allocation shares must sum to the expense; confirmed deposits and refundable amounts cannot become negative & $1{,}852$ declared signed updates, median $15$ each (range $5$--$30$) \\
\cmidrule(lr){1-4}
Validity rules & records that affect state & confirmed bookings and receipts are valid; canceled items, fully refunded payments, tentative quotes, duplicated forwards, and superseded screenshots are excluded & $1{,}019$ rules drawn from $756$ distinct status labels ($404$ invalid, $352$ valid) \\
\cmidrule(lr){1-4}
Noise channels & state-preserving distractors & small talk, stickers, hypothetical suggestions, aliases, duplicated messages, and receipts backfilled out of chronological order & $442$ distractor templates; out-of-order rate $0.02$--$0.05$ \\
\cmidrule(lr){1-4}
Difficulty controls & scalable generation parameters & number of members, blocks, and events; invalid-record ratio; backfill distance; entity interleaving; counterfactual depth & five tiers spanning ${\sim}7$ to ${\sim}1{,}200$ events per sample; invalid rate $0.10$--$0.14$ \\
\cmidrule(lr){1-4}
Question operators & requested state views & one of $159$ operators: running totals, Top-$N$ spenders, windowed aggregation, invalid-record auditing, comparison, or counterfactual removal & $159$ operators in seven families under four groups \\
\cmidrule(lr){1-4}
Answer program & executable supervision & replay valid events, apply the selected operator to the resulting ledger, and emit both the reference answer and its independently checkable atomic checklist & $117$ accepted simulators; $35{,}100$ training examples and $585$ held-out evaluation instances \\
\bottomrule
\end{tabularx}
\end{table}

\paragraph{Instance controls.}
The final question acts as a controlled probe of the reconstructed state. We define $159$ distinct operators organized into seven families under four broader groups covering ranking and comparison, aggregation and profiling, audit and counterfactual reasoning, and localization. Different operators induce different access patterns over the same event history. Ranking operators compare reconstructed values across entities, windowed operators restrict computation to a selected portion of the trajectory, and counterfactual operators remove designated events before replaying the remainder. Pairing the same underlying process with different operators therefore changes what the model must recover without changing the evolution semantics.

The difficulty setting determines how much information evolution the model must process. We define five ordered tiers that jointly scale the number of blocks, events per block, invalid-record rate, backfill distance, entity interleaving, and counterfactual depth. The resulting instances span approximately $7$ to $1{,}200$ events per sample. Question operators and difficulty settings thus play complementary roles. Question operators determine the requested state view, whereas difficulty settings determine how difficult it is to reconstruct that view. Several operator families require multi-step numerical or relational computation over the reconstructed state~\citep{cobbe2021gsm8k,wei2022cot}. Appendix~\ref{app:opbank} summarizes the final-question bank and provides representative operators and questions.

\paragraph{Multi-turn context construction.}
Given a confirmed parameter set, each simulator execution is rendered as a fixed natural-language, multi-turn context. The system message states the task-specific tracking and validity conventions, and each user turn contributes a new block of events. Some turns also contain an intermediate question followed by a code-derived stage-wise response generated by the simulator. These intermediate question--response pairs are included as fixed parts of the context to illustrate the expected question--answer format. They are neither generated by the evaluated model nor scored during evaluation. At evaluation time, the model receives the complete rendered context and is asked to answer only the final question, which requires integrating the full event history.

The renderer interleaves valid records with invalidated, out-of-order, and state-preserving distractor records without exposing the running state directly. The model must therefore determine which records apply and reconstruct the state required by the final question. Each final question is paired with the code-derived reference answer and atomic checklist defined in Section~\ref{sec:simulator-synthesis}. The checklist decomposes the expected answer into independently verifiable claims, enabling response evaluation without requiring the judge to replay the complete history. Appendix~\ref{app:example} provides worked examples of the resulting format, including one event history queried through all seven question families.

\begin{figure}[tp]
\centering
\includegraphics[width=\textwidth]{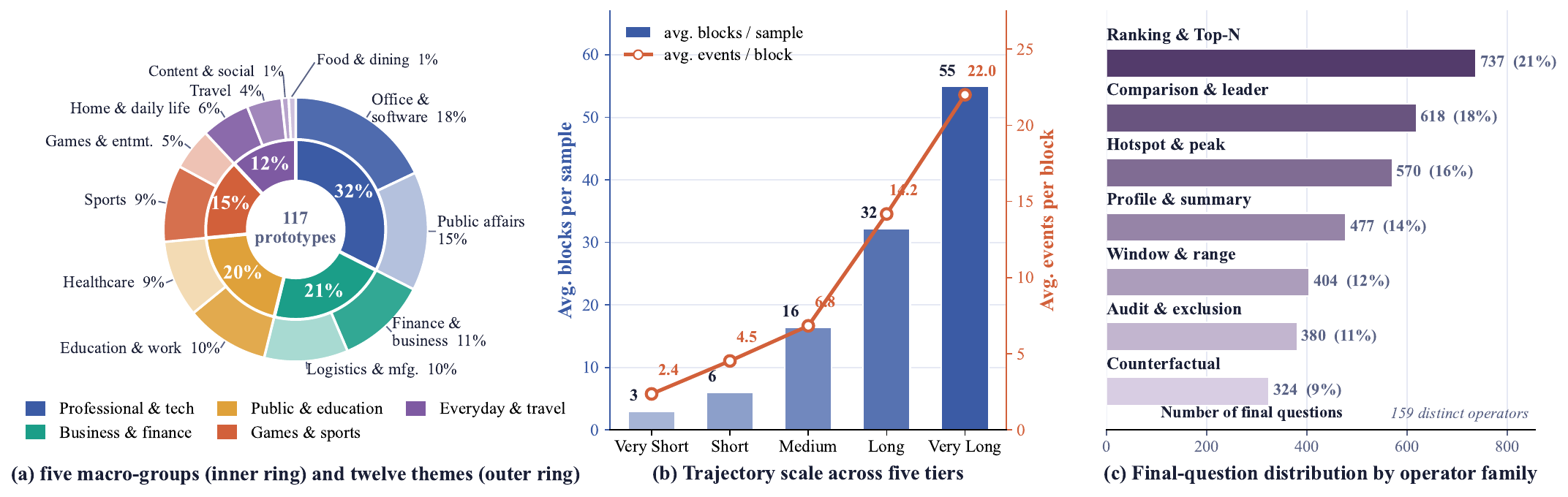}
\caption{Task coverage, trajectory scale, and final-question distribution in \evolve{}.}
\label{fig:stats}
\end{figure}

\paragraph{Corpus construction.}
We apply this controlled generation process to $117$ accepted simulators spanning $12$ themes under five macro-groups. For each simulator and difficulty tier, we sample $60$ training instances and one held-out instance. This produces approximately $35{,}100$ training examples and $585$ generated held-out examples, with $7{,}020$ training examples and $117$ held-out examples in each tier. Figure~\ref{fig:stats} summarizes the resulting corpus in terms of task-theme coverage, context scale, and final-question distribution.

% ===========================================================================
\section{Experiments}
% ===========================================================================

We evaluate \evolve{} in its two intended roles and further analyze the failure modes exposed by the evaluation suite. We first describe the evaluation protocol for the held-out data in Section~\ref{sec:eval-setup}, followed by a comparison of $14$ frontier and open-source LLMs in Section~\ref{sec:eval}. We then use the training subsets for continued training and evaluate transfer to eight independently constructed out-of-distribution benchmarks in Section~\ref{sec:train}. Finally, Section~\ref{sec:error} examines the main error patterns in failed responses.

\subsection{Evaluation Setup}
\label{sec:eval-setup}

At evaluation time, each model receives the complete rendered multi-turn dialogue, including simulator-generated intermediate question--response pairs from earlier turns. These pairs are fixed in the input and serve as in-context examples of the expected question--answer format. Only the model's response to the final user question is scored. Each response is evaluated against its code-derived atomic checklist. The checklist judge receives only the final user question, the candidate response, and the checklist. It does not see the preceding dialogue, and its role is limited to determining whether the specified atomic claims are satisfied. We use \texttt{gpt-oss-120b} with its default inference parameters as the checklist judge; Appendix~\ref{app:judge-template} provides the exact grading template.

Question sampling is controlled separately from sequence length. Within each task prototype, final questions are sampled uniformly from its available question operators at every tier, keeping the operator mix and the associated question-difficulty distribution broadly balanced across tiers. We evaluate five ordered sequence-length tiers: \texttt{very\_short}, \texttt{short}, \texttt{medium}, \texttt{long}, and \texttt{very\_long}. Longer sequences contain more events and require models to integrate more state transitions, increasing state-reconstruction and reasoning complexity and thereby making the instances more difficult. Reasoning-effort variants are identified explicitly in the model names in Table~\ref{tab:eval}; all other inference parameters use the same settings supplied by the model provider. Every input fits within the evaluated model's supported context window and is processed without truncation.

We report performance using two complementary metrics, \textbf{pass@5} and \textbf{avg@5}. A response passes only when it satisfies every item in its atomic checklist, in which case it receives a binary value of $1$; all other responses receive $0$. The \textbf{pass@5} metric is the proportion of instances for which at least one of five independently generated responses receives $1$. The \textbf{avg@5} metric is the average of these five binary values, computed across all instances. Pass@5 indicates whether a model can solve an instance at least once, whereas avg@5 reflects response-level reliability.

\subsection{Evaluation with Frontier and Open-Source LLMs}
\label{sec:eval}

Table~\ref{tab:eval} compares frontier and open-source models across five difficulty tiers. GPT-5.5-xhigh achieves the best overall performance, with a pass@5 of $82.7$ and an avg@5 of $72.8$. Performance declines sharply as the tiers jointly increase trajectory scale and evolution complexity. At each tier, we compute the median of the avg@5 scores across all evaluated models. This value falls from $71.2$ on \texttt{very\_short} to $67.8$ on \texttt{short}, $50.5$ on \texttt{medium}, $21.9$ on \texttt{long}, and $11.3$ on \texttt{very\_long}. The decline becomes especially pronounced on the two hardest tiers. Six models score below an avg@5 of $10$ on \texttt{very\_long}. Only the three GPT-5.5 variants exceed an avg@5 of $30$, while DeepSeek-V4-Preview-Pro is close at $29.7$. These results show that \evolve{} covers a broad and unsaturated difficulty range. Its easier tiers remain relatively tractable, while its hardest tiers continue to challenge even the strongest models.

% ===== Table: frontier and open models on EvolveScaler =====
\begin{table}[tp]
\centering
\caption{\textbf{Frontier and open-source models on \evolve{}} (\%; per sequence-length tier and overall,
each as pass@5 / avg@5). Best per column in \textbf{bold}, second best \underline{underlined};
rows are sorted by overall avg@5. VS / S / M / L / VL $=$ \texttt{very\_short} / \texttt{short} /
\texttt{medium} / \texttt{long} / \texttt{very\_long}. The bottom two rows summarize
each column across all $14$ models.}
\label{tab:eval}
\resizebox{\textwidth}{!}{%
\begin{tabular}{@{}lcccccccccccc@{}}
\toprule
\multirow{2}{*}{\textbf{Model}}
 & \multicolumn{2}{c}{\textbf{VS}} & \multicolumn{2}{c}{\textbf{S}}
 & \multicolumn{2}{c}{\textbf{M}} & \multicolumn{2}{c}{\textbf{L}}
 & \multicolumn{2}{c}{\textbf{VL}} & \multicolumn{2}{c}{\textbf{Average}} \\
\cmidrule(lr){2-3}\cmidrule(lr){4-5}\cmidrule(lr){6-7}\cmidrule(lr){8-9}\cmidrule(lr){10-11}\cmidrule(lr){12-13}
 & p@5 & a@5 & p@5 & a@5 & p@5 & a@5 & p@5 & a@5 & p@5 & a@5 & p@5 & a@5 \\
\midrule
GPT-5.5-xhigh & 83.8 & \underline{78.5} & 83.8 & \textbf{76.4} & \textbf{79.5} & \textbf{76.6} & \textbf{85.5} & \textbf{73.2} & \textbf{81.2} & \textbf{59.3} & \textbf{82.7} & \textbf{72.8} \\
GPT-5.5-high & \textbf{87.2} & \textbf{81.7} & 82.9 & 75.2 & \textbf{79.5} & 73.8 & \underline{83.8} & \underline{69.7} & \underline{75.2} & \underline{47.5} & \underline{81.7} & \underline{69.6} \\
GPT-5.5-medium & 83.8 & 77.9 & \underline{87.2} & \underline{75.7} & \textbf{79.5} & \underline{74.2} & 77.8 & 62.7 & 59.8 & 37.9 & 77.6 & 65.7 \\
Gemini-3.1-Pro & 84.6 & 74.5 & 81.2 & 67.5 & 75.2 & 59.8 & 61.5 & 42.2 & 47.0 & 27.5 & 69.9 & 54.3 \\
Hy-3-high & 84.6 & 72.1 & \textbf{88.0} & 74.4 & \underline{78.6} & 63.6 & 73.5 & 39.7 & 34.2 & 19.8 & 71.8 & 53.9 \\
DeepSeek-V4-Preview-Pro & 83.8 & 66.2 & 82.9 & 62.2 & 74.4 & 52.5 & 67.5 & 40.0 & 56.4 & 29.7 & 73.0 & 50.1 \\
GLM-5.2 & \underline{86.3} & 76.6 & 82.9 & 70.6 & 75.2 & 48.7 & 45.3 & 21.5 & 26.5 & 12.5 & 63.2 & 46.0 \\
Qwen3.5-Plus-Thinking & 80.3 & 62.7 & 75.2 & 64.8 & 72.6 & 51.6 & 35.0 & 22.2 & 16.2 & 10.1 & 55.9 & 42.3 \\
GLM-5.1 & \underline{86.3} & 74.9 & 83.8 & 68.0 & 64.1 & 38.8 & 25.6 & 13.7 & 16.2 & 9.6 & 55.2 & 41.0 \\
Hy3-Preview-high & \underline{86.3} & 70.3 & 80.3 & 70.4 & 68.4 & 42.2 & 26.5 & 14.4 & 10.3 & 5.1 & 54.4 & 40.5 \\
Doubao-2.0-Pro-high & 71.8 & 59.1 & 71.8 & 63.4 & 65.0 & 49.4 & 28.2 & 16.8 & 12.8 & 9.1 & 49.9 & 39.6 \\
Doubao-1.8-high & 76.1 & 59.0 & 75.2 & 65.6 & 52.1 & 35.0 & 17.1 & 11.5 & 12.8 & 8.4 & 46.7 & 35.9 \\
Doubao-2.0-Pro-medium & 75.2 & 60.3 & 71.8 & 60.7 & 37.6 & 27.5 & 16.2 & 9.4 & 9.4 & 4.8 & 42.1 & 32.5 \\
Doubao-1.8-medium & 73.5 & 60.0 & 71.8 & 61.5 & 35.9 & 21.2 & 12.8 & 7.5 & 11.1 & 6.2 & 41.0 & 31.3 \\
\midrule
\textbf{Average} & 81.7 & 69.6 & 79.9 & 68.3 & 67.0 & 51.1 & 46.9 & 31.8 & 33.5 & 20.5 & 61.8 & 48.2 \\
\textbf{Median} & 83.8 & 71.2 & 82.1 & 67.8 & 73.5 & 50.5 & 40.2 & 21.9 & 21.4 & 11.3 & 59.6 & 44.1 \\
\bottomrule
\end{tabular}}
\end{table}

The harder tiers also provide substantially greater separation among models. The range of avg@5 scores spans only $59.0$--$81.7$ on \texttt{very\_short}, but widens to $7.5$--$73.2$ on \texttt{long} and $4.8$--$59.3$ on \texttt{very\_long}. GPT-5.5-xhigh and GLM-5.1, for example, differ by only $3.6$ points on \texttt{very\_short}, where they score $78.5$ and $74.9$, but by $49.7$ points on \texttt{very\_long}, where they score $59.3$ and $9.6$. The harder tiers therefore do more than lower overall performance. They expose substantial differences in models' ability to reconstruct evolving state; these differences remain largely hidden on easier instances. This increased separation gives \evolve{} greater diagnostic value for comparing frontier and open-source models.

The harder tiers further reveal model behavior that is not captured by an overall score alone. On \texttt{very\_long}, GPT-5.5-medium achieves a pass@5 of $59.8$ and an avg@5 of $37.9$. The corresponding scores are $56.4$ and $29.7$ for DeepSeek-V4-Preview-Pro, $34.2$ and $19.8$ for Hy-3-high, and $26.5$ and $12.5$ for GLM-5.2. These differences distinguish models that can occasionally produce a successful response under repeated sampling from those that do so reliably. Within the GPT-5.5 family, the three reasoning-effort variants remain close on the first three tiers but diverge on the harder tiers. Avg@5 increases from $62.7$ to $69.7$ and $73.2$ on \texttt{long}, and from $37.9$ to $47.5$ and $59.3$ on \texttt{very\_long}, as reasoning effort increases from medium to high and xhigh. Together, these results show that the harder tiers reveal both differences in response reliability and, within the GPT-5.5 family, gains from increased reasoning effort that remain less visible on easier instances.

\subsection{Training Generalization of \evolve{}}
\label{sec:train}

We continue training an internal A3B model using the GRPO algorithm on mixed-tier \evolve{} examples and evaluate transfer on eight independently constructed benchmarks whose instances are absent from the training data. These benchmarks include MultiChallenge~\citep{deshpande2025multichallenge}, InverseIF~\citep{zhang2025inverseifeval}, IF-Bench~\citep{pyatkin2025ifbench}, MARS~\citep{yang2025marsbench}, CL-Life~\citep{dou2026cllife}, CL-Bench~\citep{dou2026cl}, MRCR~\citep{vodrahalli2024michelangelo}, and AA-LCR~\citep{artificialanalysis2025aalcr}. Table~\ref{tab:train} compares the base model with continued-training settings using $200$, $2{,}000$, and $6{,}000$ examples.

% ===== Table: out-of-distribution generalization of continued training =====
\begin{table}[tp]
\centering
\caption{\textbf{Out-of-distribution generalization of continued training on \evolve{}} (\%).
\textbf{An internal A3B model} undergoes continued training on $200$\,/\,$2{,}000$\,/\,$6{,}000$ \evolve{} samples and is
evaluated on eight held-out benchmarks. Best per column in \textbf{bold}; Avg.\ is the unweighted mean across benchmarks, and the last row is the gain over the base.}
\label{tab:train}
\resizebox{\textwidth}{!}{%
\begin{tabular}{@{}lccccccccc@{}}
\toprule
\textbf{Training setting} & \textbf{MultiChallenge} & \textbf{InverseIF} & \textbf{IF-Bench} & \textbf{MARS} & \textbf{CL-Life} & \textbf{CL-Bench} & \textbf{MRCR} & \textbf{AA-LCR} & \textbf{Avg.} \\
\midrule
Internal A3B model (base)              & 43.22 & 51.93 & 43.95 & 49.88 & 3.46 & 7.15 & 15.09 & 50.67 & 33.17 \\
\midrule
\,\,+ \evolve{} ($200$)        & 46.28 & 53.51 & 43.06 & 55.09 & 4.32 & 8.21 & 19.31 & 54.99 & 35.60 \\
\,\,+ \evolve{} ($2{,}000$)    & 48.23 & 55.04 & \textbf{46.53} & 57.23 & 4.57 & \textbf{9.48} & 21.74 & 56.80 & 37.45 \\
\,\,+ \evolve{} ($6{,}000$)    & \textbf{50.55} & \textbf{56.82} & 46.05 & \textbf{59.38} & \textbf{5.31} & \textbf{9.48} & \textbf{22.68} & \textbf{57.08} & \textbf{38.42} \\
\midrule
$\Delta$ ($6{,}000-$base)
 & \textcolor{teal!65!black}{+7.33} & \textcolor{teal!65!black}{+4.89} & \textcolor{teal!65!black}{+2.10}
 & \textcolor{teal!65!black}{+9.50} & \textcolor{teal!65!black}{+1.85} & \textcolor{teal!65!black}{+2.33}
 & \textcolor{teal!65!black}{+7.59} & \textcolor{teal!65!black}{+6.41} & \textcolor{teal!65!black}{+5.25} \\
\bottomrule
\end{tabular}}
\end{table}

The average score across the eight benchmarks increases monotonically with training scale. It rises from $33.17$ for the base model to $35.60$, $37.45$, and $38.42$, yielding a total gain of $5.25$ points at $6{,}000$ examples. The first $200$ examples improve the average by $2.43$ points, followed by additional gains of $1.85$ and $0.97$ points. At $6{,}000$ examples, performance exceeds the base model on all eight benchmarks. The largest gains occur on MARS ($+9.50$), MRCR ($+7.59$), MultiChallenge ($+7.33$), and AA-LCR ($+6.41$). The remaining benchmarks also improve, with gains of $4.89$ on InverseIF, $2.33$ on CL-Bench, $2.10$ on IF-Bench, and $1.85$ on CL-Life. Because none of these benchmark instances is included in the continued-training data, the broad gains provide evidence that the benefits of \evolve{} supervision extend beyond its own task format.

\begin{figure}[tp]
\centering
\includegraphics[width=\textwidth]{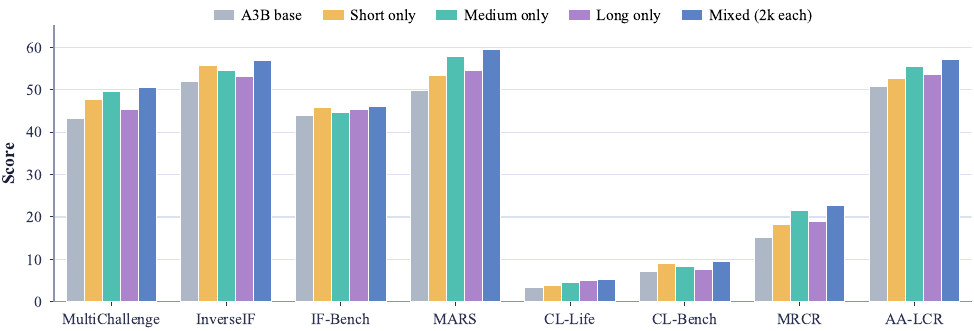}
\caption{\textbf{Training-length composition at a fixed $6{,}000$-example budget.} Each single-tier setting uses $6{,}000$ examples from one tier, whereas the mixed setting uses $2{,}000$ examples from each of the short, medium, and long tiers. The mixed-length setting achieves the strongest performance on all eight benchmarks.}
\label{fig:train-composition}
\end{figure}

To examine the effect of length composition independently of the total training budget, Figure~\ref{fig:train-composition} shows the base model and compares three single-tier training sets with a mixed-length set under the same $6{,}000$-example budget. The mixed set contains $2{,}000$ examples from each of the short, medium, and long tiers. It achieves the strongest performance on all eight benchmarks, while the single-tier settings exhibit benchmark-dependent trade-offs. This pattern suggests that exposure to multiple trajectory scales provides complementary supervision.

\subsection{Error Analysis}
\label{sec:error}

Pass@5 and avg@5 summarize whether a model succeeds, but they do not show where a failed response breaks down. We therefore manually inspect failed responses across models and question operators. Each semantic failure receives one primary label according to the first atomic checklist item it fails to satisfy. \emph{Invalid-Record Handling} marks cases in which the model constructs the wrong set of valid records. \emph{Retrieval \& Counting} covers cases in which relevant records are omitted or irrelevant records are included. \emph{Net-Change Aggregation} applies when the selected records are appropriate but their signed effects are combined incorrectly. \emph{Ranking \& Tie-Breaking} captures errors in ordering or deterministic tie resolution. \emph{Comparison Verdict} covers cases in which the underlying values are correct but the winner or margin is wrong or missing. We do not add a generic calculation category because nearly every \evolve{} query requires computation. Formatting errors are negligible; blank outputs and refusals are excluded. Table~\ref{tab:error} reports the distribution of these primary labels for ten models.

% ===== Table: distribution of semantic error types =====
\begin{table}[tp]
\centering
\footnotesize
\caption{\textbf{Distribution of semantic error types across representative models}
(\% of semantic failures; blank/refusal outputs excluded). Each failure is assigned one
primary type according to the first atomic checklist item it fails to satisfy.}
\label{tab:error}
\setlength{\tabcolsep}{3pt}
\renewcommand{\arraystretch}{1.08}
\begin{tabular}{@{}p{3.55cm}p{2.2cm}p{2.2cm}p{2.2cm}p{2.2cm}p{2.2cm}@{}}
\toprule
\multicolumn{1}{c}{\textbf{Model}}
& \multicolumn{1}{c}{\shortstack{\textbf{Invalid-Record}\\\textbf{Handling (\%)}}}
& \multicolumn{1}{c}{\shortstack{\textbf{Retrieval \&}\\\textbf{Counting (\%)}}}
& \multicolumn{1}{c}{\shortstack{\textbf{Net-Change}\\\textbf{Aggregation}\\\textbf{(\%)}}}
& \multicolumn{1}{c}{\shortstack{\textbf{Ranking \&}\\\textbf{Tie-Breaking}\\\textbf{(\%)}}}
& \multicolumn{1}{c}{\shortstack{\textbf{Comparison}\\\textbf{Verdict (\%)}}} \\
\midrule
GPT-5.5-xhigh          & \scorebarerr{4.5}{4.5} & \scorebarerr{18.4}{18.4} & \scorebarerr{46.2}{46.2} & \scorebarerr{29.8}{29.8} & \scorebarerr{1.1}{1.1} \\
Gemini-3.1-Pro         & \scorebarerr{2.0}{2.0} & \scorebarerr{16.8}{16.8} & \scorebarerr{50.7}{50.7} & \scorebarerr{21.9}{21.9} & \scorebarerr{8.5}{8.5} \\
Hy-3-high            & \scorebarerr{4.7}{4.7} & \scorebarerr{20.1}{20.1} & \scorebarerr{49.6}{49.6} & \scorebarerr{22.5}{22.5} & \scorebarerr{3.1}{3.1} \\
DeepSeek-V4-Preview-Pro        & \scorebarerr{2.9}{2.9} & \scorebarerr{19.3}{19.3} & \scorebarerr{48.6}{48.6} & \scorebarerr{20.7}{20.7} & \scorebarerr{8.5}{8.5} \\
GLM-5.2                & \scorebarerr{5.0}{5.0} & \scorebarerr{21.1}{21.1} & \scorebarerr{50.3}{50.3} & \scorebarerr{22.4}{22.4} & \scorebarerr{1.3}{1.3} \\
Qwen3.5-Plus-Thinking  & \scorebarerr{4.4}{4.4} & \scorebarerr{17.3}{17.3} & \scorebarerr{55.6}{55.6} & \scorebarerr{19.2}{19.2} & \scorebarerr{3.5}{3.5} \\
GLM-5.1                & \scorebarerr{5.5}{5.5} & \scorebarerr{17.7}{17.7} & \scorebarerr{53.8}{53.8} & \scorebarerr{20.3}{20.3} & \scorebarerr{2.8}{2.8} \\
Hy3-Preview-high       & \scorebarerr{4.8}{4.8} & \scorebarerr{19.7}{19.7} & \scorebarerr{53.3}{53.3} & \scorebarerr{19.6}{19.6} & \scorebarerr{2.5}{2.5} \\
Doubao-2.0-Pro-high    & \scorebarerr{4.3}{4.3} & \scorebarerr{17.3}{17.3} & \scorebarerr{52.9}{52.9} & \scorebarerr{19.4}{19.4} & \scorebarerr{6.1}{6.1} \\
Doubao-1.8-high        & \scorebarerr{4.9}{4.9} & \scorebarerr{17.2}{17.2} & \scorebarerr{55.9}{55.9} & \scorebarerr{19.3}{19.3} & \scorebarerr{2.6}{2.6} \\
\bottomrule
\end{tabular}
\end{table}

Net-Change Aggregation is the largest error category for every reported model and accounts for roughly half of the labeled semantic failures. Its share is lowest for GPT-5.5-xhigh at $46.2\%$ and highest for Doubao-1.8-high at $55.9\%$. Retrieval \& Counting falls within the narrower range of $16.8\%$--$21.1\%$. Together, these two categories constitute at least $64.6\%$ and as much as $73.1\%$ of the labeled failures. Most observed errors therefore surface when models select the records that contribute to the reconstructed state or combine the signed effects of those records. Ranking \& Tie-Breaking is also substantial. It accounts for at least $19.2\%$ of the labeled failures and reaches $29.8\%$ for GPT-5.5-xhigh. Notably, GPT-5.5-xhigh has both the lowest share of Net-Change Aggregation errors and the highest share of Ranking \& Tie-Breaking errors. This combination shows that deterministic ordering and tie resolution remain prominent in the residual errors of the strongest model. By contrast, Invalid-Record Handling never exceeds $5.5\%$, and Comparison Verdict does not exceed $8.5\%$. Because each response receives only the label associated with its first violated checklist claim, these proportions should not be interpreted as independent error rates. An early validity error may later appear as a retrieval, counting, or aggregation failure. Even with this limitation, the analysis shows that code-derived atomic checklists provide a structured view of where failures surface during state reconstruction and downstream computation. This gives \evolve{} diagnostic value beyond pass@5 and avg@5 alone and reveals where information evolution remains most difficult.

\section{Conclusion}

We introduced \evolve{}, a code-driven framework for synthesizing data that targets information evolution in long, multi-turn interactions. In these settings, the context is not merely a document to read but an event history whose ordered updates must be replayed to recover the state required by the final question. \evolve{} begins with human-authored operational specifications that define state transitions, record validity, difficulty controls, and answer logic. A strong LLM compiles each specification into an executable simulator, but the replayed program state, rather than the LLM, determines the supervision. Simulator execution produces natural-language event histories, while deterministic replay yields reference answers and atomic checklists. This separation allows varied natural-language realizations while keeping the underlying semantics and ground truth under executable control. The resulting resource comprises $117$ task prototypes and $159$ final-question operators across five difficulty tiers.

As an evaluation suite, \evolve{} remains challenging for frontier and open-source models. As trajectory scale and evolution complexity increase, the median of the models' avg@5 scores falls from $71.2$ on \texttt{very\_short} to $11.3$ on \texttt{very\_long}, and the harder tiers reveal substantially larger performance differences among models. As a training source, \evolve{} supports continued training of an internal A3B model on $6{,}000$ examples using the GRPO algorithm, improving performance on all eight independently constructed out-of-distribution benchmarks and raising their average score by $5.25$ points. The code-derived atomic checklists further show that many labeled failures surface during record selection, net-change aggregation, and tie-aware ranking. Together, these findings show that reasoning over evolving information remains difficult for current models and that executable synthesis can support both diagnostic evaluation and transferable training supervision. The current framework focuses on discrete, programmatically specified state transitions. Extending it to partially observed, continuous, and multimodal settings remains an important direction for future work.

\FloatBarrier

% ===========================================================================
\bibliographystyle{iclr2025_conference}
\bibliography{iclr2025_conference}

% ===========================================================================
\newpage
\appendix
% ===========================================================================

\section{Task Domains}
\label{app:domains}
The $117$ \evolve{} prototypes span $12$ themes under five macro-groups, deliberately covering registers in which
information genuinely accumulates and mutates over time. Table~\ref{tab:app_domains} lists \emph{all} $117$
prototypes, grouped exactly as in Figure~\ref{fig:stats}(a). The distribution is deliberately uneven: process-intensive
registers, where records are filed, revised, and retracted as a matter of routine, carry more prototypes than
registers in which information mostly accumulates without being corrected.

{\small
\setlength{\tabcolsep}{4pt}
\renewcommand{\arraystretch}{1.25}
\begin{longtable}{@{}>{\raggedright\arraybackslash}p{2.45cm}>{\raggedright\arraybackslash}p{2.65cm}>{\raggedright\arraybackslash}p{10.3cm}@{}}
\caption{\textbf{The $117$ task prototypes of \evolve{},} grouped into $12$ themes under five macro-groups.
Counts in parentheses give the number of prototypes.}\label{tab:app_domains}\\
\toprule
\textbf{Macro-group} & \textbf{Theme} & \textbf{Prototypes} \\
\midrule
\endfirsthead
\toprule
\textbf{Macro-group} & \textbf{Theme} & \textbf{Prototypes} \\
\midrule
\endhead
\bottomrule
\endfoot
\textbf{Professional \& tech}\newline\textit{(38)} & Office \& software \textit{(21)} & Annotation QC log; API spec revisions; CI/CD deployment log; Cloud cost optimization; Code review thread; Data pipeline incident; Dataset curation log; Design feedback stream; Energy grid event stream; IoT device logs; Localization changes; Network routing changes; PRD revision flow; Project versioned editing; Prompt iteration log; Release notes revisions; Sensor stream monitoring; Server outage timeline; Slide deck revision; Taxonomy building flow; UX research notes \\
 & Public affairs \textit{(17)} & Archival timeline notes; Committee voting thread; Compliance audit flow; Contract negotiation; Court docket updates; Detective case board; Diplomacy negotiation log; Election count updates; Emergency coordination chat; Fire response timeline; Historical policy process; Legal discovery timeline; Legislative amendments; Museum collection log; Online moderation queue; Security incident response; Treaty clause tracking \\
\midrule
\textbf{Business \& finance}\newline\textit{(25)} & Finance \& business \textit{(13)} & Ad campaign budgeting; Auction bidding log; Budget revision thread; Expense reconciliation; Fundraising donor log; Insurance claims flow; Invoice matching queue; Loan portfolio updates; Procurement bid process; Real estate listing updates; Sales CRM pipeline; Subscription churn log; Trading ledger updates \\
 & Logistics \& mfg. \textit{(12)} & Farm irrigation log; Flight disruption updates; Food delivery dispatch; Logistics dispatch board; Manufacturing line log; Route planning log; Shipment route tracking; Survival resource log; Traffic count stream; Train timetable changes; Warehouse inventory ops; Weather station updates \\
\midrule
\textbf{Public \& education}\newline\textit{(23)} & Education \& work \textit{(12)} & Classroom forum planning; Classroom gradebook; Debate score flow; Grant review queue; Hiring pipeline flow; Job application pipeline; Lab experiment log; Library lending records; Literature review coding; Parent--teacher thread; Peer review process; Study group quiz \\
 & Healthcare \textit{(11)} & Ambulance dispatch log; Clinic queue management; Clinical trial visit log; Hospital triage queue; Lab results tracker; Medication inventory; Nutrition log updates; Public health case trace; Rehab session log; Symptom tracker stream; Vaccination booking flow \\
\midrule
\textbf{Games \& sports}\newline\textit{(17)} & Sports \textit{(11)} & Sim racing telemetry; Badminton rally; Baseball inning; Basketball play-by-play; Handball attack; Ice hockey shift; Rugby phase; Soccer timeline; Table tennis points; Tennis point log; Volleyball rally \\
 & Games \& entmt. \textit{(6)} & Board game moves; Card game turns; Chess-like match log; Escape room clues; Puzzle hunt progress; Roleplay RPG combat \\
\midrule
\textbf{Everyday \& travel}\newline\textit{(14)} & Home \& daily life \textit{(7)} & Apartment chores chat; Conference scheduling; Customer chat preferences; Group chat scheduling; Multiplayer strategy chat; Spreadsheet ledger rows; Survey collection stream \\
 & Travel \textit{(5)} & City walking tour; Conference travel plan; Hotel booking thread; Itinerary constraints; Travel group discussion \\
 & Content \& social \textit{(1)} & Editorial board chat \\
 & Food \& dining \textit{(1)} & Restaurant planning chat \\
\end{longtable}
}

\section{The Final-Question Bank}
\label{app:opbank}
Table~\ref{tab:app_ops} details the bank of $159$ final-question operators, grouped into seven families under four groups. Each row gives example operator names and a representative (paraphrased) question; shares are calculated over $3{,}510$ final questions randomly sampled from the training set.

{\small
\setlength{\tabcolsep}{4pt}
\renewcommand{\arraystretch}{1.25}
% Snake_case operator names are unhyphenatable, so let them break after "_".
\renewcommand{\_}{\textunderscore\allowbreak}
\begin{longtable}{@{}>{\raggedright\arraybackslash}p{2.15cm}>{\raggedright\arraybackslash}p{2.5cm}>{\raggedright\arraybackslash}p{5.0cm}>{\raggedright\arraybackslash}p{5.35cm}@{}}
\caption{\textbf{The final-question bank.} There are seven families under four groups in total. Example operator names and a representative question of each family are shown in the table.}\label{tab:app_ops}\\
\toprule
\textbf{Group} & \textbf{Family (share)} & \textbf{Example operators} & \textbf{Example question (paraphrased)} \\
\midrule
\endfirsthead
\toprule
\textbf{Group} & \textbf{Family (share)} & \textbf{Example operators} & \textbf{Example question (paraphrased)} \\
\midrule
\endhead
\bottomrule
\endfoot
Ranking \& comparison & Ranking / Top-$N$ ($21\%$) & \texttt{event\_type\_actor\_leaderboard}, \texttt{component\_rank\_top\_n\_metric} & Counting only confirmed records, rank the top-3 accounts by spend---what are the values? \\
 & Comparison \& leader ($18\%$) & \texttt{actor\_target\_dual\_metric}, \texttt{actor\_best\_target\_dual} & Compare accounts A and B on remaining budget and spend; which has the higher remaining budget? \\
\midrule
Aggregation \& profiling & Profile \& summary ($14\%$) & \texttt{actor\_full\_trajectory}, \texttt{actor\_event\_type\_breakdown} & Summarize actor X's valid trajectory: current metrics and the latest valid record ID? \\
 & Window \& range ($12\%$) & \texttt{event\_type\_window\_summary}, \texttt{recent\_window\_actor\_leader} & Over batches 1--5 only, what is the net change in pending-invoices and remaining budget? \\
\midrule
Audit \& counterfactual & Audit \& exclusion ($11\%$) & \texttt{invalid\_exclusion\_with\_total}, \texttt{event\_type\_total\_recount} & Which records are invalid? After excluding them, what is the overall valid alert count? \\
 & Counterfactual ($9\%$) & \texttt{counterfactual\_remove\_actor}, \texttt{counterfactual\_remove\_target} & Counterfactual: dropping all valid records of account X, what is the remaining valid pending-invoice total? \\
\midrule
Localization & Hotspot \& peak ($16\%$) & \texttt{actor\_target\_hotspot}, \texttt{busiest\_turn\_for\_event\_type} & Which project has the most valid ``budget top-up'' records, and what is its net spend change? \\
\end{longtable}
}

\section{Checklist-Judge Template}
\label{app:judge-template}

We use \texttt{gpt-oss-120b} with its default inference parameters as the checklist judge. A response is assigned $1$ only when every checklist item is satisfied, and $0$ otherwise. For each instance, avg@5 is the mean of these binary values across five independently generated responses; pass@5 is $1$ if at least one of the five responses receives $1$, and $0$ otherwise. The following grading template is supplied verbatim for every candidate response:

\begin{lstlisting}[
  basicstyle=\ttfamily\scriptsize,
  breaklines=true,
  breakatwhitespace=false,
  columns=fullflexible,
  keepspaces=true,
  showstringspaces=false
]
From now on, your role is a rigorous instruction-following grader. Your task is to grade the student's answer precisely according to the <Scoring Checklist>.

## Scoring Principle
Every requirement in the <Scoring Checklist> is equally important and carries the same weight. When determining the final score, you must consider all requirements in the checklist jointly. A student answer that violates multiple requirements should receive a lower score, while a student answer that satisfies all requirements should receive a higher score.

## Grading Procedure
You must strictly follow the steps below and must not skip any part.

### Step 1: Analyze the reference criteria
* List all explicit requirements in the <Scoring Checklist> one by one (including format, content, quantity, order, etc.).
* Identify the implicit requirements in the <Scoring Checklist> (such as language style or logical structure).
* Define concrete evaluation standards for each requirement (for example: ``must include X'', ``must not exceed Y'').

### Step 2: Check the student answer against each requirement
* For each requirement in the <Scoring Checklist>, verify one by one whether the student answer fully satisfies it.

### Step 3: Self-reflection
Before giving the final score, you must conduct the following checks:
* Completeness check: Have all requirements in the <Scoring Checklist> been reviewed without omission?
* Strictness check: Did you adhere to the standard of ``fully satisfied'' without relaxing the requirements based on subjective judgment?
* Consistency check: Are the scoring rationale and the final score logically consistent?
* Objectivity check: Is the judgment based on objective evidence rather than subjective speculation?

## Output Format Requirements
Your output must contain exactly three parts: [Scoring Rationale], [Requirement Satisfaction Status List], and [Score]. Do not output any additional content. The output format must be exactly as follows:

<begin_of_Scoring_Rationale>xxx<end_of_Scoring_Rationale>
<begin_of_Requirement_Satisfaction_Status_List>[x_1, x_2, ..., x_i, ..., x_n] (where n is the total number of requirements in the <Scoring Checklist>, and x_i indicates whether the student answer satisfies the i-th requirement; each x_i must be either 0 or 1.)<end_of_Requirement_Satisfaction_Status_List>
<begin_of_Score>x points (The score must be an integer from 0 to 10. Please assign an overall quality score between 0 and 10 based on the degree to which the student answer satisfies the requirements. If all requirements are violated, assign 0. If all requirements are satisfied, assign 10.)<end_of_Score>

## I hope you can fulfill the role of a grading teacher well, because this is very important to my work. If you do well, I will give you an appropriate reward. Otherwise, I may impose an appropriate penalty. The formal question is as follows:

<Question>:
{question}

<Scoring Checklist>:
{checklist}

<Student Answer>:
{response}
\end{lstlisting}

\section{Worked Evolution Examples}
\label{app:example}
This appendix gives three abridged samples drawn from different registers, together covering all seven
final-question families. In every case, the answer does not appear in any single line of the log; it can only be recovered by
replaying the trace and dropping the invalid records. Example~2 additionally shows the \emph{same} trace answered
under all seven families, which is the mechanism by which one prototype yields many distinct reasoning demands.

\subsection{Example 1: an editorial pitch group (audit \& exclusion)}
Table~\ref{tab:app_example} shows one abridged \texttt{short}-tier sample. Only formally confirmed messages count;
old screenshots, hearsay, mock drafts, tests, and retracted messages are invalid.

{\small
\setlength{\tabcolsep}{0pt}
\renewcommand{\arraystretch}{1.2}
\begin{longtable}{@{}p{\linewidth}@{}}
\caption{\textbf{A worked evolution example} (editorial pitch group; \texttt{short} tier, abridged and translated).}\label{tab:app_example}\\
\toprule
\endfirsthead
\toprule
\endhead
\bottomrule
\endfoot
\textbf{System.} Track an editorial pitch group. Answer only from formally confirmed messages, post-meeting syncs, and schedule entries; pre-meeting brainstorm, old schedule screenshots, unverified hearsay, mock samples, tests, and retracted messages do \emph{not} count. \\*[3pt]
\midrule
\textbf{Turn 1 (user).} Newly confirmed:
\;$\bullet$ \texttt{T01-E01} (Day\,1 09:00, \emph{old screenshot}): pitch ``indie-caf\'e delivery'' passed; passed-pitch $+1$. \emph{[superseded]}
\;$\bullet$ \texttt{T01-E02} (09:07, schedule sync): Tang flags a risk on ``old-bookstore night log''; risk-flags $+1$.
\;$\bullet$ \texttt{T01-E03} (09:14, schedule sync): a to-do added---``subway breakfast-cart'' still needs materials; pending-material $+1$.
\emph{\;Q: over all valid ``risk report'' records, what is the change in risk-flags, and who contributed most?} \;\textbf{Assistant:} $+1$; top contributor Tang. \\*[3pt]
\midrule
\textbf{Turn 2 (user).} Newly synced:
\;$\bullet$ \texttt{T02-E01} (09:21): materials completed for ``dorm-AC dispute''; pending-material $-1$.
\;$\bullet$ \texttt{T02-E02} (09:28): materials completed for ``old-bookstore''; pending-material $-1$, risk-flags $-1$.
\;$\bullet$ \texttt{T02-E03} (09:35, \emph{old screenshot}): pitch ``subway breakfast-cart'' passed. \emph{[superseded]}
\;$\bullet$ \texttt{T02-E04} (09:42, post-meeting confirm): ``old-bookstore'' scheduled to Han; scheduled $+1$.
\emph{\;Q (same as before).} \;\textbf{Assistant:} $+1$; top contributor Tang. \\*[3pt]
\midrule
\textbf{Turn 3 (user).}
\;$\bullet$ \texttt{T03-E01} (09:49): Tang reports a new execution risk on ``old-bookstore''; risk-flags $+1$.
\;$\bullet$ \texttt{T03-E02} (09:56): push back the ``subway breakfast-cart'' release; scheduled $-1$.
\;$\bullet$ \texttt{T03-E03} (10:03, confirm): ``indie-caf\'e'' slotted into the next issue; scheduled $+1$.
\emph{\;Q: total valid risk-flag count so far?} \;\textbf{Assistant:} $1$. \\*[3pt]
\midrule
\textbf{Final question} (\emph{audit \& exclusion}). Which records are invalid? After excluding them, what is the overall valid \emph{pending-material} count? \\*[3pt]
\midrule
\textbf{Checklist (code-derived).} \;(1)~invalid records $=$ \texttt{T01-E01}, \texttt{T02-E03}; \;(2)~overall valid pending-material count $=-1$. \\*[2pt]
\textbf{Code-verified answer.} Invalid: \texttt{T01-E01}, \texttt{T02-E03} (old screenshots). Excluding them, the pending-material count is $\mathbf{-1}$: $+1$ (\texttt{T01-E03}) $-1$ (\texttt{T02-E01}) $-1$ (\texttt{T02-E02}); the two screenshots touch only passed-pitch and are dropped. \\
\end{longtable}
}

\subsection{Example 2: a warehouse shift board, answered seven ways}
Table~\ref{tab:app_example2} shows a \texttt{medium}-tier trace from warehouse inventory operations. Four crews act on
four SKU-locations with five tracked metrics. Table~\ref{tab:app_sevenq} then poses one final question from each of
the seven families \emph{for this same trace}. The context never changes; only the requested view of the
reconstructed state does.

{\small
\setlength{\tabcolsep}{0pt}
\renewcommand{\arraystretch}{1.2}
\begin{longtable}{@{}p{\linewidth}@{}}
\caption{\textbf{A warehouse shift board} (warehouse inventory operations; \texttt{medium} tier, abridged and translated).}\label{tab:app_example2}\\
\toprule
\endfirsthead
\toprule
\endhead
\bottomrule
\endfoot
\textbf{System.} Track a warehouse shift board across receiving, put-away, picking, and replenishment. Only WMS-posted, floor-re-checked, supervisor-confirmed, and stock-control-confirmed records count; drill tasks, retracted scans, old count drafts, and test waves do \emph{not}. Running metrics: inbound units, outbound units, relocated units, pending discrepancy, exceptions. \\*[3pt]
\midrule
\textbf{Wave 1 (user).}
\;$\bullet$ \texttt{W1-R1} (08:10, WMS posted): East Dock Receiving books in \texttt{A1-03} bottled water; inbound $+120$.
\;$\bullet$ \texttt{W1-R2} (08:18, \emph{drill task}): Night Picking ships \texttt{B1-05} jasmine tea; outbound $+80$. \emph{[drill]}
\;$\bullet$ \texttt{W1-R3} (08:25, floor re-checked): West Rack Put-away replenishes \texttt{A2-11} tissue packs; inbound $+45$, relocated $+45$.
\;$\bullet$ \texttt{W1-R4} (08:33, WMS posted): Night Picking picks \texttt{A1-03} bottled water; outbound $+60$.
\emph{\;Q: over all valid receipts, what is the change in inbound units?} \;\textbf{Assistant:} $+165$. \\*[3pt]
\midrule
\textbf{Wave 2 (user).}
\;$\bullet$ \texttt{W2-R1} (09:02, supervisor confirmed): Cold-chain Replenishment books in \texttt{D4-08} detergent; inbound $+90$.
\;$\bullet$ \texttt{W2-R2} (09:09, \emph{old count draft}): East Dock Receiving logs a count gap on \texttt{A1-03}; pending discrepancy $+12$. \emph{[superseded]}
\;$\bullet$ \texttt{W2-R3} (09:15, WMS posted): East Dock Receiving finds a count gap on \texttt{A2-11}; pending discrepancy $+18$, exceptions $+1$.
\;$\bullet$ \texttt{W2-R4} (09:24, stock-control confirmed): Night Picking ships \texttt{B1-05} jasmine tea; outbound $+75$.
\;$\bullet$ \texttt{W2-R5} (09:31, floor re-checked): West Rack Put-away relocates \texttt{A1-03}; relocated $+30$.
\emph{\;Q (same as before).} \;\textbf{Assistant:} $+255$. \\*[3pt]
\midrule
\textbf{Wave 3 (user).}
\;$\bullet$ \texttt{W3-R1} (10:05, WMS posted): East Dock Receiving clears the \texttt{A2-11} gap; pending discrepancy $-18$.
\;$\bullet$ \texttt{W3-R2} (10:12, \emph{retracted scan}): Cold-chain Replenishment ships \texttt{D4-08}; outbound $+40$. \emph{[retracted]}
\;$\bullet$ \texttt{W3-R3} (10:20, supervisor confirmed): Night Picking recovers a mis-pick on \texttt{A1-03}; outbound $-15$, exceptions $+1$.
\;$\bullet$ \texttt{W3-R4} (10:28, WMS posted): Cold-chain Replenishment ships \texttt{D4-08} detergent; outbound $+55$.
\;$\bullet$ \texttt{W3-R5} (10:36, floor re-checked): West Rack Put-away replenishes \texttt{B1-05} jasmine tea; inbound $+60$, relocated $+60$.
\emph{\;Q: total valid exceptions so far?} \;\textbf{Assistant:} $2$. \\*[3pt]
\midrule
\textbf{Hidden state after replay} (valid records only; never shown to the model). Inbound $315$, outbound $175$, relocated $135$, pending discrepancy $0$, exceptions $2$. Invalid: \texttt{W1-R2}, \texttt{W2-R2}, \texttt{W3-R2}. \\
\end{longtable}
}

{\small
\setlength{\tabcolsep}{4pt}
\renewcommand{\arraystretch}{1.3}
\begin{longtable}{@{}>{\raggedright\arraybackslash}p{2.6cm}>{\raggedright\arraybackslash}p{6.2cm}>{\raggedright\arraybackslash}p{6.6cm}@{}}
\caption{\textbf{One trace, seven final questions.} Each row applies a different operator family to the
\emph{identical} log in Table~\ref{tab:app_example2}; every answer is computed by replay, not judged.}\label{tab:app_sevenq}\\
\toprule
\textbf{Family} & \textbf{Final question (paraphrased)} & \textbf{Code-verified answer} \\
\midrule
\endfirsthead
\toprule
\textbf{Family} & \textbf{Final question (paraphrased)} & \textbf{Code-verified answer} \\
\midrule
\endhead
\bottomrule
\endfoot
Ranking / Top-$N$ & Counting only valid records, rank the top-3 crews by inbound units. & East Dock Receiving $120$; West Rack Put-away $105$; Cold-chain Replenishment $90$. \\
\midrule
Comparison \& leader & Compare West Rack Put-away and Cold-chain Replenishment on inbound and outbound units. & West Rack leads inbound ($105$ vs.\ $90$); Cold-chain leads outbound ($55$ vs.\ $0$). \\
\midrule
Profile \& summary & Summarize Night Picking's valid trajectory and give its latest valid record. & Outbound $120$, exceptions $1$, all other metrics $0$; latest valid record \texttt{W3-R3}. \\
\midrule
Window \& range & Over waves $2$--$3$ only, what is the net change in outbound units and pending discrepancy? & Outbound $+115$ ($+75-15+55$); pending discrepancy $0$ ($+18$ then $-18$). \\
\midrule
Audit \& exclusion & Which records are invalid? After excluding them, what is the total outbound? & Invalid: \texttt{W1-R2} (drill), \texttt{W2-R2} (old draft), \texttt{W3-R2} (retracted). Total outbound $175$. \\
\midrule
Counterfactual & Dropping all valid records of Night Picking, what is the remaining outbound? & $55$: only \texttt{W3-R4} survives, since Night Picking contributed $+60+75-15=120$ of the $175$. \\
\midrule
Hotspot \& peak & Which SKU-location carries the most valid records, and what is its inbound? & \texttt{A1-03} bottled water, with $4$ valid records; its inbound is $120$. \\
\end{longtable}
}

\subsection{Example 3: a clinical-trial follow-up log (profile and comparison)}
Table~\ref{tab:app_example3} shows a \texttt{short}-tier trace from a different register, in which the sources of
invalid records are training drills and screenshot replays rather than stale screenshots or drills on a shop floor.

{\small
\setlength{\tabcolsep}{0pt}
\renewcommand{\arraystretch}{1.2}
\begin{longtable}{@{}p{\linewidth}@{}}
\caption{\textbf{A clinical-trial follow-up log} (clinical trial visit log; \texttt{short} tier, abridged and translated).}\label{tab:app_example3}\\
\toprule
\endfirsthead
\toprule
\endhead
\bottomrule
\endfoot
\textbf{System.} Track a clinical-trial follow-up log across visit stations. Only signed-and-entered, investigator-confirmed, EDC-locked, and monitor-reviewed records count; old drafts, training drills, verbal transcriptions, retracted records, and screenshot replays do \emph{not}. Running metrics: completed items, pending re-review, risk flags, waiting minutes. \\*[3pt]
\midrule
\textbf{Visit day 1 (user).}
\;$\bullet$ \texttt{V1-R1} (signed \& entered): subject Han completes the vitals room; completed items $+1$.
\;$\bullet$ \texttt{V1-R2} (\emph{training drill}): subject Zhou completes the blood draw; completed items $+1$. \emph{[drill]}
\;$\bullet$ \texttt{V1-R3} (investigator confirmed): a risk is reported for Han at the blood draw; risk flags $+1$, pending re-review $+1$.
\;$\bullet$ \texttt{V1-R4} (signed \& entered): subject Chen waits longer at the check-in desk; waiting $+25$ min.
\emph{\;Q: how many valid completed items so far?} \;\textbf{Assistant:} $1$. \\*[3pt]
\midrule
\textbf{Visit day 2 (user).}
\;$\bullet$ \texttt{V2-R1} (EDC locked): Han's blood-draw data passes review; pending re-review $-1$.
\;$\bullet$ \texttt{V2-R2} (\emph{screenshot replay}): Han completes the ECG room; completed items $+1$. \emph{[replay]}
\;$\bullet$ \texttt{V2-R3} (monitor review passed): subject Zhou completes the ECG room; completed items $+1$.
\;$\bullet$ \texttt{V2-R4} (signed \& entered): a protocol deviation is filed for Chen at the investigator office; risk flags $+1$, pending re-review $+1$.
\;$\bullet$ \texttt{V2-R5} (investigator confirmed): subject Zhou waits longer at the vitals room; waiting $+15$ min. \\*[3pt]
\midrule
\textbf{Final question A} (\emph{profile \& summary}). Summarize subject Han's valid trajectory: current metrics and the latest valid record. \\*[2pt]
\textbf{Code-verified answer.} Completed items $1$, risk flags $1$, pending re-review $\mathbf{0}$ (raised by \texttt{V1-R3}, cleared by \texttt{V2-R1}), waiting $0$ min; latest valid record \texttt{V2-R1}. The ECG entry \texttt{V2-R2} is a screenshot replay and does not count. \\*[3pt]
\midrule
\textbf{Final question B} (\emph{comparison \& leader}). Compare subjects Chen and Zhou on completed items and risk flags. \\*[2pt]
\textbf{Code-verified answer.} Zhou leads completed items ($1$ vs.\ $0$; Zhou's day-1 blood draw was a training drill and is dropped, but \texttt{V2-R3} counts). Chen leads risk flags ($1$ vs.\ $0$) and is the only subject with an open pending re-review. \\
\end{longtable}
}

% ===========================================================================
% Parked: "Planned further analyses" table. It lived in tables/further.tex and
% was never \input by the paper. Kept here (commented) so the draft is not lost;
% delete this block or uncomment it to bring the table into the appendix.
% ===========================================================================
% \begin{table}[tp]
% \centering
% \small
% \setlength{\tabcolsep}{6pt}
% \renewcommand{\arraystretch}{1.25}
% \caption{\textbf{Planned further analyses.} Each isolates one factor while holding
% the rest fixed, exploiting \evolve{}'s code-level control over difficulty.}
% \label{tab:further}
% \begin{tabular}{@{}>{\raggedright\arraybackslash}p{3.2cm}>{\raggedright\arraybackslash}p{5.1cm}>{\raggedright\arraybackslash}p{5.4cm}@{}}
% \toprule
% \textbf{Analysis} & \textbf{Question} & \textbf{Setup} \\
% \midrule
% Difficulty-knob ablation & how much does each factor move scores? & fix the theme; scale event count, invalid ratio, and counterfactual depth one at a time \\
% Failure-mode attribution & which abilities are weakest? & break frontier-model errors down by the final-question operator families \\
% Data-mixture scaling & how do mixture and size affect transfer? & vary the length-tier composition and total size of the training data \\
% \bottomrule
% \end{tabular}
% \end{table}

\end{document}